\documentclass[11pt, a4paper]{article}
\pdfoutput=1
\usepackage{lrec}
\usepackage{vntex}
\usepackage[english]{babel}
\usepackage{graphicx}
\usepackage{tabularx}

\usepackage{fixltx2e}

\usepackage{times}

\usepackage[hidelinks]{hyperref}

\title{A Fast and Accurate Vietnames{e}\ {W}ord Segmenter}

\name{Dat Quoc Nguyen$^1$, Dai Quoc Nguyen$^2$, Thanh Vu$^3$, Mark Dras$^4$, Mark Johnson$^4$}

\address{$^1$The University of Melbourne, Australia \\
{\tt dqnguyen@unimelb.edu.au}\\
         $^2$Deakin University, Australia \\
         {\tt dai.nguyen@deakin.edu.au}\\
         $^3$Newcastle University, United Kingdom\\
{\tt thanh.vu@newcastle.ac.uk} \\
$^4$Macquarie University, Australia \\
{\tt\{mark.dras, mark.johnson\}@mq.edu.au}\\}

\abstract{
We propose a novel approach to Vietnamese word segmentation.  Our approach is based on the Single Classification Ripple Down Rules methodology \cite{ComptonJ90}, where rules are stored in an exception structure and new rules are only added to correct segmentation errors given by existing rules. Experimental results on the benchmark Vietnamese treebank show that our approach outperforms previous state-of-the-art approaches JVnSegmenter, vnTokenizer,  DongDu and UETsegmenter in terms of both accuracy and performance speed. Our code is open-source and available at: \url{https://github.com/datquocnguyen/RDRsegmenter}. \\
\newline  }

\begin{document}

\maketitleabstract

\section{Introduction}\label{sec:intro}

\noindent Word segmentation is referred to as an important first
step for Vietnamese NLP tasks \cite{Dien2001,Ha2003,DucCong2016}.
Unlike English,  white space is a weak indicator of word boundaries in Vietnamese because when written, it is 
 also  used to separate syllables that constitute  words. 
For example, a written text ``thuế thu nhập  cá nhân'' (individual\textsubscript{cá\_nhân} income\textsubscript{thu\_nhập} tax\textsubscript{thuế}) consisting of 5 syllables forms a two-word phrase ``thuế\_thu\_nhập  cá\_nhân.''\footnote{In the traditional  underscore-based representation in the Vietnamese word segmentation task \protect\cite{nguyen-EtAl:2009:LAW-III}, white space is only used to separate words
while underscore is used to separate syllables inside a word.} More specifically, about 85\% of Vietnamese word types are composed of at least two syllables  and 80\%+ of syllable types are words by themselves \cite{DinhQuangThang2008,Le2008}, thus creating  challenges in Vietnamese word segmentation \cite{nguyenNM2012}.

Many approaches are proposed for the Vietnamese word segmentation task. \newcite{Le2008}, \newcite{Pham2009} and \newcite{Tran2012}  applied  the maximum matching strategy \cite{NanYuan91} to generate all possible segmentations for each input sentence; then to select the best segmentation, \newcite{Le2008} and \newcite{Tran2012} used  n-gram language models while \newcite{Pham2009} employed part-of-speech (POS) information from an external POS tagger. 
In addition, \newcite{Y06-1028},   
\newcite{Dinh2006} and \newcite{Tran2010} considered this segmentation task as a sequence labeling task, using either a linear-chain CRF, SVM or MaxEnt model to assign each syllable a segmentation tag such as B (Begin of a word) or I (Inside of a word). Another promising approach is joint  word segmentation and POS tagging  \cite{Takahashi2016,NguyenVNDJ-ALTA-2017}, which assigns a combined segmentation and POS tag to each syllable. Furthermore, \newcite{LuuTA2012}, \newcite{Liu2014} and \newcite{NguyenL2016} proposed  methods based on pointwise prediction \cite{NEUBIG10.408}, where a binary classifier is trained to identify whether or not there is a word boundary  between two syllables.

In this paper, we propose a novel method to  Vietnamese word segmentation. Our method automatically constructs a Single Classification Ripple Down Rules  (\textit{SCRDR}) tree \cite{ComptonJ90} to correct wrong segmentations given by a longest matching-based word segmenter.   
On the benchmark Vietnamese treebank \cite{nguyen-EtAl:2009:LAW-III}, experimental results  show that our method obtains better accuracy and performance speed than the previous state-of-the-art methods JVnSegmenter \cite{Y06-1028}, vnTokenizer \cite{Le2008},  DongDu \cite{LuuTA2012} and UETsegmenter  \cite{NguyenL2016}.

\begin{figure*}[ht]
\centering
\includegraphics[width=17cm]{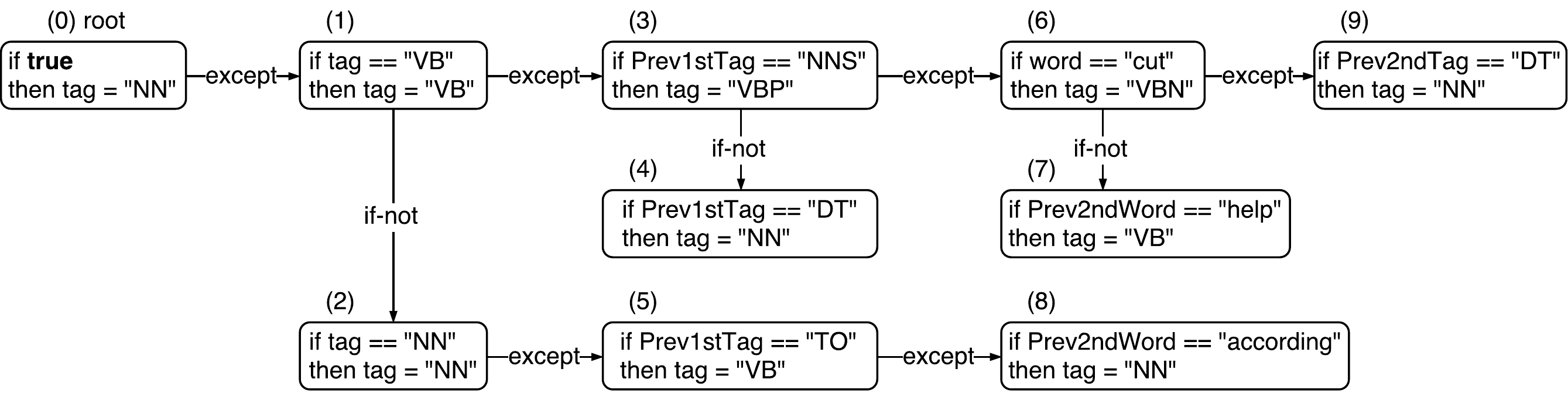} 
\caption{An illustration of a SCRDR tree for POS tagging.  This figure is adapted from \protect\newcite{NguyenNPP_AICom2015}.}
\label{fig:kbscrdr}
\end{figure*}

\section{SCRDR methodology}
\label{sec:scrdr}

\noindent This section  gives a brief introduction of the SCRDR methodology \cite{ComptonJ88,ComptonJ90,RichardsD09}. 
A SCRDR tree is a binary tree with only  two  unique types of edges  ``except'' and ``if-not'', where  every node  is associated with a \textit{rule} in a form of ``\textit{if} \textbf{condition} \textit{then} \textbf{conclusion}.'' To ensure that the tree always produces a conclusion, the rule at its root (default) node  has a trivial condition which is always satisfied.

Each case to be evaluated starts at the root node and  ripples down as follows: (i) If the case satisfies the condition of a current node's rule, the case is then passed on to the current node's ``except'' child if this ``except'' child exists. (ii) Otherwise, if the case does not satisfy the condition, it is then passed on to the current node's ``if-not'' child. So, the conclusion returned by the tree is the conclusion of the last satisfied rule in the evaluation path to a leaf node.

For example, Figure \ref{fig:kbscrdr} illustrates a SCRDR tree for POS tagging. Let us consider a concrete case ``\textit{as/IN investors/NNS {anticipate/VB} a/DT recovery/NN}'' where  ``\textit{anticipate}'' and ``\textit{VB}'' is the current considered pair of word and its initial POS tag. 
Because this case satisfies the conditions of the rules at nodes (0), (1) and (3), it is  passed on to  node (6) using the ``except'' edge. The case does not satisfy the condition of the rule at node (6), thus it is passed on to node (7) using the ``if-not'' edge. 
Finally, the case does not satisfy the condition of the rule at  the leaf node (7). So, the rule at node (3)---the last satisfied rule in the the evaluation path (0)-(1)-(3)-(6)-(7)---concludes ``\textit{VBP}'' should be the POS tag of the word ``\textit{anticipate}'' instead of the initial POS tag ``\textit{VB}.''

To correct a wrong conclusion returned for a given case, a new node containing a new \textit{exception} rule may be attached to  the last node in the evaluation path. If the last node's rule is the last satisfied rule  given the case, the new node is added as its child with the ``except'' edge; otherwise, the new node is attached with the ``if-not'' edge. 

 SCRDR has been successfully applied in NLP tasks for temporal relation extraction \cite{PhamSB:2006}, word lemmatization \cite{Plisson2008}, POS tagging \cite{Xu:2010,NguyenNPP2011,NguyenEACL2014NPP,NguyenNPP_AICom2015}, named entity recognition \cite{NguyenBD2012} and question answering \cite{NguyenRANLP2011NP,Nguyen2013,NguyenNP_SWJ}.  
The works by \newcite{Plisson2008}, \newcite{NguyenNPP2011}, \newcite{NguyenEACL2014NPP} and \newcite{NguyenNPP_AICom2015}  build the tree automatically, while others manually  construct the  tree.

\begin{figure}[t]
\centering
\includegraphics[width=8cm]{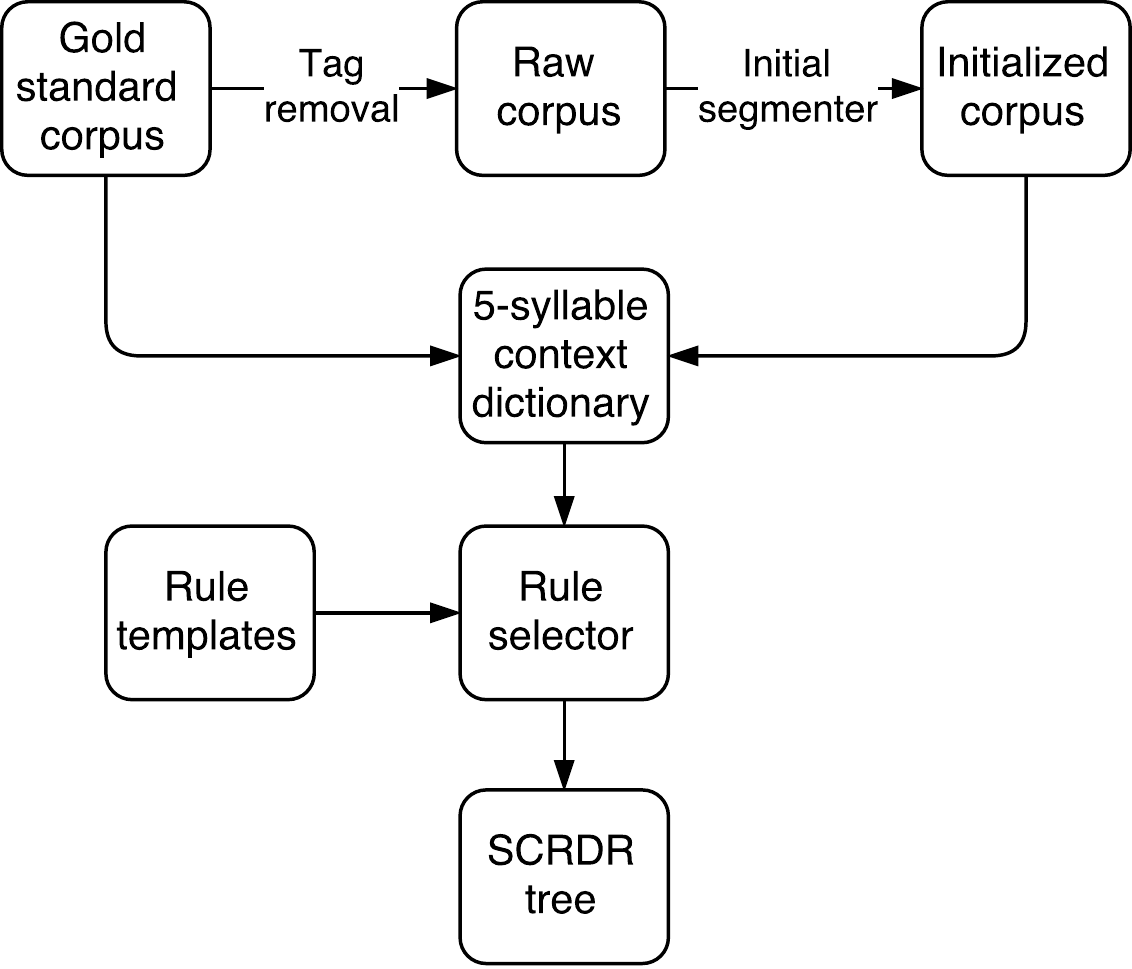} 
\caption{Diagram of our   approach.}
\label{fig:diagram}
\end{figure}

\section{Our approach}

\noindent This section describes our new error-driven approach to automatically construct  a SCRDR tree to correct wrong segmentations produced by an initial word segmenter.

Following \newcite{Y06-1028} and \newcite{Tran2010},   we also formalize the word segmentation problem as a sequence labeling task.  In particular, each syllable is labeled by either segmentation tag B (Begin of a word) or I (Inside of a word). 
As a result, our approach can be viewed as an extension to word segmentation of  
the automatic SCRDR approach for POS tagging \cite{NguyenEACL2014NPP,NguyenNPP_AICom2015}.  Our  learning diagram  is described in Figure \ref{fig:diagram}.

We start with an underscore-based \textit{gold standard training corpus} consisting of manually word-segmented sentences, e.g. ``thuế\_thu\_nhập  cá\_nhân'' (individual\textsubscript{cá\_nhân} income\textsubscript{thu\_nhập} tax\textsubscript{thuế}) and transform this corpus into a BI-formed representation (e.g. ``thuế/B thu/I nhập/I  cá/B nhân/I''). 
We then extract syllables to construct the \textit{raw corpus} (which does not have B and I segmentation tags, and would look like ``thuế thu nhập  cá nhân''). 

We apply an \textit{initial segmenter} on the input {raw corpus} to get the output BI-formed \textit{initialized corpus}.  For example, given the input raw text ``thuế thu nhập  cá nhân'', the initial segmenter returns  the output BI-formed initialized text ``thuế/B thu/B nhập/I  cá/B nhân/I.'' 
The {initial segmenter} in our approach is based on the longest matching strategy \cite{Poowarawan}, using a Vietnamese lexicon from  \newcite{Le2008}.

 We then compare the BI-formed gold standard corpus and the BI-formed initialized   corpus to generate a \textit{5-syllable context dictionary} $\mathcal{D}$ where each key-value pair consists of a 5-syllable window tuple as key and a gold standard tag as value. Here, each tuple captures a 5-syllable window context of a current syllable and its initial  segmentation tag B/I in a format of (Previous-2nd-syllable, Previous-2nd-tag, Previous-1st-syllable, Previous-1st-tag, syllable, tag, Next-1st-syllable, Next-1st-tag, Next-2nd-syllable, Next-2nd-tag)  from the initialized corpus,\footnote{Syllables in each tuple are all converted into a lowercase form.} while the  gold standard tag 
 is the corresponding segmentation tag of the current syllable in the gold standard corpus.   So, a wrong segmentation is when the initial segmentation tag is different from the gold standard  tag, as shown in the second row in Table \ref{tab:dict}.

\begin{table}[t]
\centering
\begin{tabular}{l|cl }
\hline 
Tuple as key & Value & \\
\hline
(\textbf{``''}, \textbf{``''}, \textbf{``''}, \textbf{``''}, \textbf{thuế}, \textbf{B}, {thu}, {B}, nhập, I) & B & $\surd$ \\
(\textbf{``''}, \textbf{``''}, thuế, B, \textbf{thu}, \textbf{B}, nhập, I,  cá, B) & I  & $\mathsf{X}$ \\
(thuế, B, thu, B, \textbf{nhập}, \textbf{I},  cá, B, nhân, I) & I  & $\surd$ \\
(thu, B, nhập, I,  \textbf{cá}, \textbf{B}, nhân, I, \textbf{``''}, \textbf{``''}) & B & $\surd$ \\
(nhập, I,  {cá}, {B}, \textbf{nhân}, \textbf{I}, \textbf{``''}, \textbf{``''}, \textbf{``''}, \textbf{``''}) & I & $\surd$ \\
\hline
\end{tabular} 
\caption{Examples of key-value pairs in the 5-syllable context dictionary $\mathcal{D}$ when comparing the BI-formed gold standard corpus ``thuế/B thu/I nhập/I  cá/B nhân/I'' and the BI-formed initialized corpus ``thuế/B thu/B nhập/I  cá/B nhân/I.'' Here, \textbf{``''} denotes an empty element in tuples. $\surd$  and $\mathsf{X}$ represent the correct and incorrect initial segmentations, respectively.} 
\label{tab:dict}
\end{table}

\begin{table}[!t]
\centering
\resizebox{8cm}{!}{
\begin{tabular}{l|l }
\hline 
syllable & s\textsubscript{-2}, s\textsubscript{-1}, s\textsubscript{0}, s\textsubscript{+1}, s\textsubscript{+2}  \\ 
\cline{2-2} 
 & (s\textsubscript{-2}, s\textsubscript{0}), (s\textsubscript{-1}, s\textsubscript{0}), (s\textsubscript{-1}, s\textsubscript{+1}),
(s\textsubscript{0}, s\textsubscript{+1}) \\
&  (s\textsubscript{0}, s\textsubscript{+2}) \\
\cline{2-2} 
 &   (s\textsubscript{-2}, s\textsubscript{-1}, s\textsubscript{0}), (s\textsubscript{-1}, s\textsubscript{0}, s\textsubscript{+1}), (s\textsubscript{0}, s\textsubscript{+1}, s\textsubscript{+2}) \\
\hline
tag & t\textsubscript{-2}, t\textsubscript{-1},  t\textsubscript{0}, t\textsubscript{+1}, t\textsubscript{+2} \\
\cline{2-2} 
 & (t\textsubscript{-2}, t\textsubscript{-1}), (t\textsubscript{-1}, t\textsubscript{+1}), (t\textsubscript{+1}, t\textsubscript{+2}) \\
\hline
syllable \& tag & (t\textsubscript{-1}, s\textsubscript{0}), (s\textsubscript{0}, t\textsubscript{+1}),  (t\textsubscript{-1}, s\textsubscript{0}, t\textsubscript{+1}),   (t\textsubscript{-2}, t\textsubscript{-1}, s\textsubscript{0}) \\&  (s\textsubscript{0}, t\textsubscript{+1}, t\textsubscript{+2})  \\
\hline
\end{tabular} 
}
\caption{Short descriptions of our rule templates. ``s'' refers to syllable and ``t'' refers to B/I segmentation label while subscripts -2, -1, 0, 1, 2 denote indices. For example, (s\textsubscript{-1}, s\textsubscript{+1}) represents the rule  template  ``IF Previous-1st-syllable == \textbf{tuple.Previous-1st-syllable} \&\& Next-1st-syllable == \textbf{tuple.Next-1st-syllable} THEN tag = \textbf{gold-standard-tag}'', where elements in \textbf{bold} are replaced by
concrete values from tuple and gold tag pairs in the 5-syllable context dictionary $\mathcal{D}$. Given  (s\textsubscript{-1}, s\textsubscript{+1}) and the second row in Table \ref{tab:dict}, we have a concrete rule  ``IF Previous-1st-syllable == \textbf{thuế} \&\& Next-1st-syllable == \textbf{nhập} THEN tag = \textbf{I}.''} 
\label{tab:templates}
\end{table}

\begin{figure}[t]
\centering
\includegraphics[width=6cm]{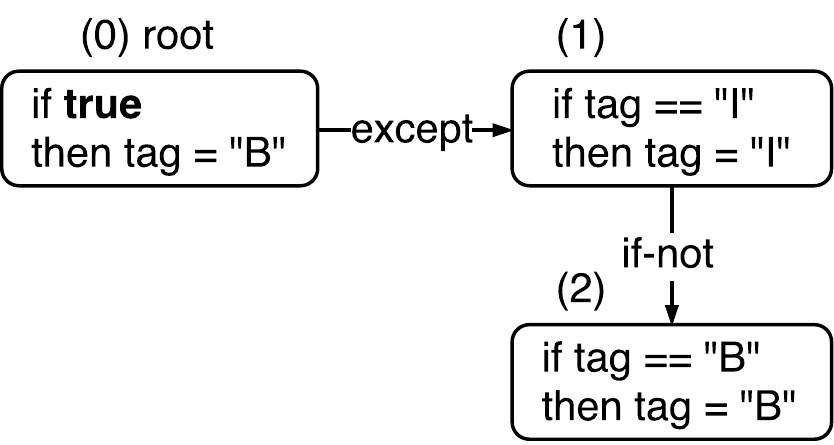} 
\caption{SCRDR tree initialization.}
\label{fig:KBinitial}
\end{figure}

Based on the 5-syllable context dictionary $\mathcal{D}$, the \textit{rule selector} selects the most suitable rules to construct the \textit{SCRDR tree}. Concrete rules are generated based on \textit{rule templates}. Table \ref{tab:templates} presents short descriptions of the rule templates. The {SCRDR tree} is  initialized with a default rule---the rule at the root node---and its two exception rules,  as shown in Figure \ref{fig:KBinitial}. 
Our learning  process to automatically add new exception rules to the  SCRDR tree is as follows:

\begin{itemize}

\item Let us consider  a node  $\mathsf{N}$ in the tree. We define a subset $\mathcal{T}_\mathsf{N}$ of the context dictionary $\mathcal{D}$ such that the rule at  $\mathsf{N}$ is the last satisfied rule  in the evaluation path for every tuple in  $\mathcal{T}_\mathsf{N}$ but  $\mathsf{N}$ returns a wrong segmentation tag. For example, given node (2) in Figure \ref{fig:KBinitial} and  $\mathcal{D}$ in Table \ref{tab:dict},  $\mathcal{T}_\mathsf{(2)}$  would contain a pair of the tuple (\textbf{``''}, \textbf{``''}, thuế, B, \textbf{thu}, \textbf{B}, nhập, I  cá, B)  and gold segmentation tag \textbf{I}  from  the second row in Table \ref{tab:dict}.
A new node containing a new exception rule must be added to the  current tree to correct the errors given by $\mathsf{N}$.\footnote{See the second last paragraph in Section \ref{sec:scrdr} for how to attach a new node to an existing SCRDR tree.}

\item The new exception rule  is selected from all concrete rules, in which these concrete rules are generated by applying the rule templates to all tuples in $\mathcal{T}_\mathsf{N}$.  The selected rule must satisfy following constraints: (1) If $\mathsf{N}$ is not one of the first three nodes in Figure \ref{fig:KBinitial}, then the selected rule's condition must not be satisfied by every tuple for which  $\mathsf{N}$ already returns a correct segmentation tag. (2) The selected rule is associated with the highest  value of the subtraction $a - b$. Here $a$ is the number of tuples in $\mathcal{T}_\mathsf{N}$ in which each tuple not only satisfies the rule's condition but also gets a correct segmentation tag given by the rule's conclusion, while $b$ is the number of tuples in $\mathcal{T}_\mathsf{N}$ in  which each tuple also satisfies the rule's condition but gets a wrong segmentation tag given by the rule's conclusion. (3) The value $a - b$ must be not smaller than a given threshold. 

\item This process is repeated
until  at any node it cannot select a new exception rule satisfying  constraints above. 

\end{itemize}

With the learned SCRDR tree, we perform word segmentation on unsegmented text as follows: The initial segmenter takes the input 
 unsegmented text to generate a BI-formed initialized text. Next, by sliding a 5-syllable window from left to right, a tuple is generated for each syllable in the initialized text; then the learned SCRDR tree takes the input tuple to return a final segmentation tag to the corresponding syllable. Finally, the output of this labeling process is converted to the traditional underscore-based representation.

\section{Experiments}

\subsection{Experimental setup}

\noindent Following \newcite{NguyenL2016}, we conduct experiments and compare the performance of our approach---which we call RDRsegmenter---with published results of other state-of-the-art approaches on the benchmark Vietnamese treebank \cite{nguyen-EtAl:2009:LAW-III}. The training set consists of 75k manually word-segmented sentences (about 23 words per sentence in average).\footnote{The data,  officially released in 2013,  is provided  for research or educational purpose by the national  project VLSP on Vietnamese language and speech processing.} The test set consists of  2120 sentences (about 31 words per sentence) in 10 files from \textit{800001.seg} to \textit{800010.seg}.\footnote{The test set was  originally released  for evaluation in the POS tagging shared task at the  VLSP 2013 workshop.} 
 We use F$_1$ score as the main evaluation metric to measure the performance of word segmentation.

Note that to determine the threshold in our RDRsegmenter, we sampled a development set of 5k sentences from the full training set and used the remaining 70k sentences for training. We found an optimal threshold value at 2 producing the highest F$_1$ score on the development set. Then we learned a SCRDR tree from  the full training set with the optimal threshold, resulting in 1447 rules in total.

\subsection{Main results}

\noindent Table \ref{tab:results} compares the Vietnamese word segmentation results of our RDRsegmenter with results reported in prior work, using the same experimental setup. 

\begin{table}[!t]
\centering
\resizebox{8cm}{!}{
\begin{tabular}{lccc}
\hline
Approach & Precision & Recall & F$_1$ \\
\hline
vnTokenizer & 96.98 & 97.69 & 97.33  \\
JVnSegmenter-Maxent & 96.60 & 97.40 & 97.00 \\
JVnSegmenter-CRFs & 96.63 & 97.49 & 97.06 \\
DongDu  & 96.35 & 97.46 & 96.90 \\
UETsegmenter & \textbf{97.51} & 98.23 & 97.87\\
\hline
Our \textbf{RDRsegmenter} & 97.46 &	\textbf{98.35}	& \textbf{97.90}\\
\hline
\end{tabular}
}
\caption{Vietnamese word segmentation results (in \%). The results of vnTokenizer, JVnSegmenter and DongDu are reported in   \protect\newcite{NguyenL2016}.}
\label{tab:results}
\end{table}

\begin{figure}[!t]
\centering
\includegraphics[width=8.25cm]{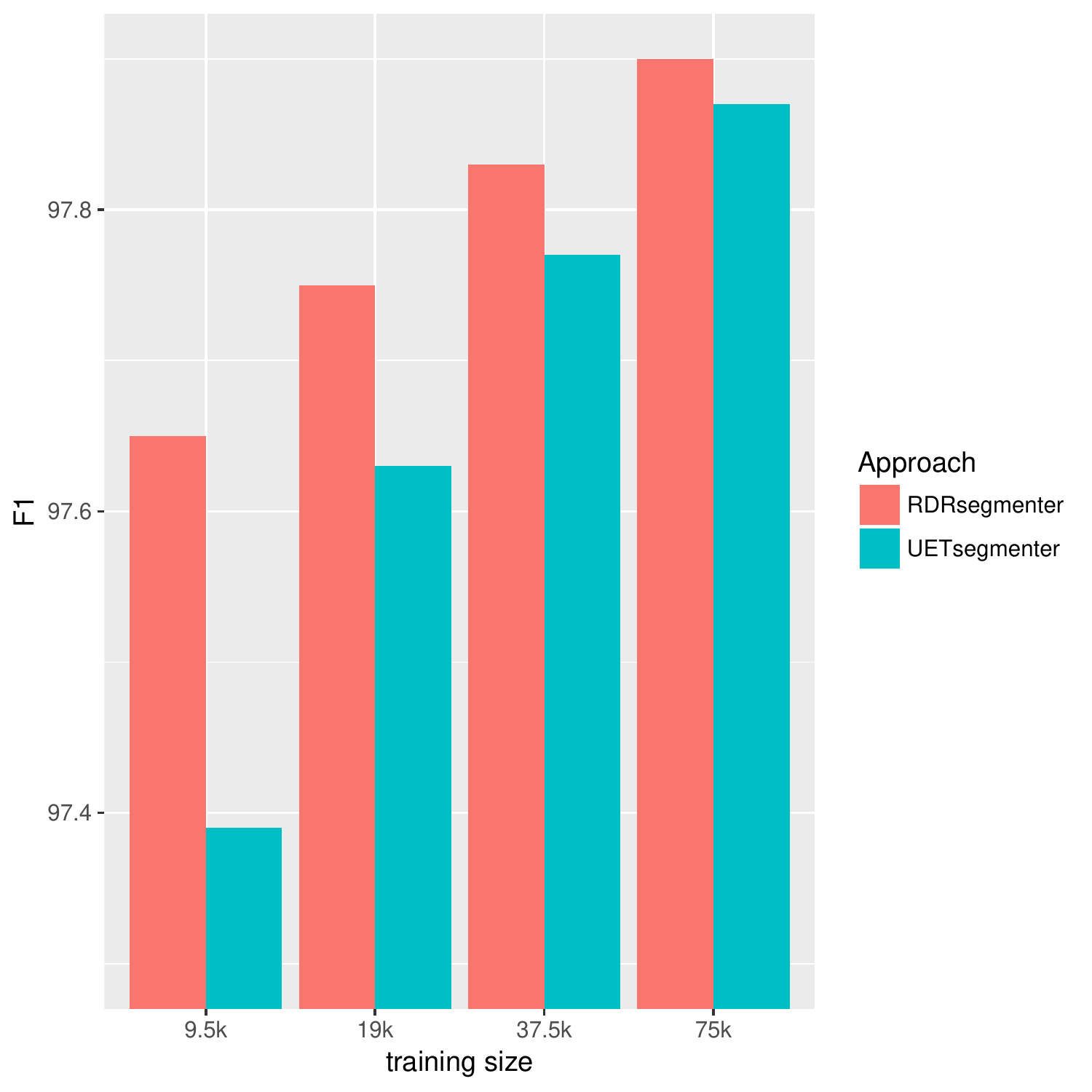} 
\caption{F$_1$ scores (in \%) when varying the training size  at 9.5k, 19k, 37.5k and full 75k sentences.}
\label{fig:f1}
\end{figure}

Table \ref{tab:results}  shows that RDRsegmenter obtains the highest F$_1$ score.
 In particular, RDRSegmenter obtains 0.5+\% higher F$_1$ than vnTokenizer \cite{Le2008} though both approaches use the same lexicon for initial segmentation. In terms of a sequence labeling task, RDRSegmenter outperforms JVnSegmenter \cite{Y06-1028} with 0.8+\% improvement. Compared with the pointwise prediction approaches DongDu \cite{LuuTA2012} and UETsegmenter \cite{NguyenL2016},  RDRsegmenter does significantly better than DongDu and somewhat better than UETsegmenter. 
 In Figure \ref{fig:f1}, we show F$_1$ scores of RDRsegmenter and UETsegmenter at different training sizes, showing that RDRsegmenter clearly improves performance in a smaller dataset scenario.

It is worth noting that on a personal computer of Intel Core i7 2.2 GHz, our  RDRsegmenter processes at a speed of 62k words per second in a single threaded implementation, which is 1.3 times faster than UETsegmenter.\footnote{We repeated the segmentation process on the test set 100 times, and then computed the averaged speed. Note that model loading time was not taken into account, in which RDRsegmenter took 50 miniseconds while UETsegmenter took 10 seconds.} In addition, \newcite{NguyenL2016} showed that UETsegmenter is faster than  vnTokenizer, JVnSegmenter and DongDu.\footnote{Evaluated on a computer of Intel Core i5-3337U 1.80GHz, \newcite{NguyenL2016} showed that JVnSegmenter, vnTokenizer, DongDu and UETsegmenter obtained performance speeds at 1k, 5k, 17k and 33k words per second, respectively. All of them are implemented in Java except DongDu which is in C++. Our  RDRsegmenter is also implemented in Java.}  So RDRsegmenter is also faster than vnTokenizer, JVnSegmenter and DongDu.

\section{Conclusion}

\noindent In this paper, we have proposed a new error-driven method
to automatically construct a Single Classification Ripple Down Rules tree for Vietnamese word segmentation. Experiments on the benchmark Vietnamese treebank show that our method obtains better accuracy and  speed than previous approaches. Our code is  available at: \url{https://github.com/datquocnguyen/RDRsegmenter}. 

Note that excluding the language-specific initial segmenter, our method generally can be viewed as a language independent approach. Here, a Vietnamese syllable is analogous to a character in other languages  such as Chinese and Japanese. So we will adapt our method to  those  languages in future work. 

\section*{Acknowledgments}
This research was partially supported by the Australian Government through the Australian Research Council's Discovery Projects funding scheme (project DP160102156). This research
was also partially supported by NICTA, funded
by the Australian Government through the Department
of Communications and the Australian Research
Council through the ICT Centre of Excellence
Program. 
This  research was done while the first author was at Macquarie University.

\section*{References}
\label{main:ref}

\bibliographystyle{lrec}
\bibliography{refs}

\begin{thebibliography}{}

\bibitem[\protect\citename{Compton and Jansen}1988]{ComptonJ88}
Compton, P. and Jansen, B.
\newblock (1988).
\newblock {Knowledge in Context: {A} Strategy for Expert System Maintenance}.
\newblock In {\em Proceedings of the 2nd Australian Joint Artificial
  Intelligence Conference}, pages 292--306.

\bibitem[\protect\citename{Compton and Jansen}1990]{ComptonJ90}
Compton, P. and Jansen, R.
\newblock (1990).
\newblock {A Philosophical Basis for Knowledge Acquisition}.
\newblock {\em Knowledge Aquisition}, 2(3):241--257.

\bibitem[\protect\citename{Dien \bgroup et al.\egroup }2001]{Dien2001}
Dien, D., Kiem, H., and Toan, N.~V.
\newblock (2001).
\newblock {Vietnamese Word Segmentation}.
\newblock In {\em Proceedings of the Sixth Natural Language Processing Pacific
  Rim Symposium}, pages 749--756.

\bibitem[\protect\citename{Dinh and Vu}2006]{Dinh2006}
Dinh, D. and Vu, T.
\newblock (2006).
\newblock {A Maximum Entropy Approach for Vietnamese Word Segmentation}.
\newblock In {\em Proceedings of the 2006 International Conference on Research,
  Innovation and Vision for the Future}, pages 248--253.

\bibitem[\protect\citename{{Duc Cong} \bgroup et al.\egroup }2016]{DucCong2016}
{Duc Cong}, S.~N., {Hung Ngo}, Q., and Jiamthapthaksin, R.
\newblock (2016).
\newblock {State-of-the-art Vietnamese word segmentation}.
\newblock In {\em Proceedings of the 2nd International Conference on Science in
  Information Technology}, pages 119--124.

\bibitem[\protect\citename{Ha}2003]{Ha2003}
Ha, L.~A.
\newblock (2003).
\newblock {A method for word segmentation in Vietnamese}.
\newblock In {\em Proceedings of Corpus Linguistics}.

\bibitem[\protect\citename{Le \bgroup et al.\egroup }2008]{Le2008}
Le, H.~P., Nguyen, T. M.~H., Roussanaly, A., and Ho, T.~V.
\newblock (2008).
\newblock {A hybrid approach to word segmentation of Vietnamese texts}.
\newblock In {\em Proceedings of the 2nd International Conference on Language
  and Automata Theory and Applications}, pages 240--249.

\bibitem[\protect\citename{Liu and Lin}2014]{Liu2014}
Liu, W. and Lin, L.
\newblock (2014).
\newblock {Probabilistic Ensemble Learning for Vietnamese Word Segmentation}.
\newblock In {\em Proceedings of the 37th International ACM SIGIR Conference on
  Research \& Development in Information Retrieval}, pages 931--934.

\bibitem[\protect\citename{Luu and Kazuhide}2012]{LuuTA2012}
Luu, T.~A. and Kazuhide, Y.
\newblock (2012).
\newblock { Ứng dụng phương pháp Pointwise vào bài toán tách từ
  cho tiếng Việt}.
\newblock https://github.com/rockkhuya/DongDu.

\bibitem[\protect\citename{NanYuan and YanBin}1991]{NanYuan91}
NanYuan, L. and YanBin, Z.
\newblock (1991).
\newblock {A Chinese word segmentation model and a Chinese word segmentation
  system PC-CWSS}.
\newblock {\em Journal of Chinese Language and Computing}, 1(1).

\bibitem[\protect\citename{Neubig and Mori}2010]{NEUBIG10.408}
Neubig, G. and Mori, S.
\newblock (2010).
\newblock {Word-based Partial Annotation for Efficient Corpus Construction}.
\newblock In {\em Proceedings of the Seventh International Conference on
  Language Resources and Evaluation}, pages 2723--2727.

\bibitem[\protect\citename{Nguyen and Le}2016]{NguyenL2016}
Nguyen, T.-P. and Le, A.-C.
\newblock (2016).
\newblock {A Hybrid Approach to Vietnamese Word Segmentation}.
\newblock In {\em Proceedings of the 2016 IEEE RIVF International Conference on
  Computing and Communication Technologies: Research, Innovation, and Vision
  for the Future}, pages 114--119.

\bibitem[\protect\citename{Nguyen and Pham}2012]{NguyenBD2012}
Nguyen, D.~B. and Pham, S.~B.
\newblock (2012).
\newblock {Ripple Down Rules for Vietnamese Named Entity Recognition}.
\newblock In {\em Proceedings of the 4th international conference on
  Computational Collective Intelligence: Technologies and Applications - Volume
  Part I}, pages 354--363.

\bibitem[\protect\citename{Nguyen \bgroup et al.\egroup }2006]{Y06-1028}
Nguyen, C.-T., Nguyen, T.-K., Phan, X.-H., Nguyen, L.-M., and Ha, Q.-T.
\newblock (2006).
\newblock {Vietnamese Word Segmentation with CRFs and SVMs: An Investigation}.
\newblock In {\em Proceedings of the 20th Pacific Asia Conference on Language,
  Information and Computation}, pages 215--222.

\bibitem[\protect\citename{Nguyen \bgroup et al.\egroup
  }2009]{nguyen-EtAl:2009:LAW-III}
Nguyen, P.~T., Vu, X.~L., Nguyen, T. M.~H., Nguyen, V.~H., and Le, H.~P.
\newblock (2009).
\newblock {Building a Large Syntactically-Annotated Corpus of Vietnamese}.
\newblock In {\em Proceedings of the Third Linguistic Annotation Workshop},
  pages 182--185.

\bibitem[\protect\citename{Nguyen \bgroup et al.\egroup
  }2011a]{NguyenRANLP2011NP}
Nguyen, D.~Q., Nguyen, D.~Q., and Pham, S.~B.
\newblock (2011a).
\newblock {Systematic Knowledge Acquisition for Question Analysis}.
\newblock In {\em Proceedings of the 8th International Conference on Recent
  Advances in Natural Language Processing}, pages 406--412.

\bibitem[\protect\citename{Nguyen \bgroup et al.\egroup }2011b]{NguyenNPP2011}
Nguyen, D.~Q., Nguyen, D.~Q., Pham, S.~B., and Pham, D.~D.
\newblock (2011b).
\newblock {Ripple Down Rules for Part-of-Speech Tagging}.
\newblock In {\em Proceedings of the 12th International Conference on
  Intelligent Text Processing and Computational Linguistics - Volume Part I},
  pages 190--201.

\bibitem[\protect\citename{Nguyen \bgroup et al.\egroup }2012]{nguyenNM2012}
Nguyen, Q.~T., Nguyen, N.~L., and Miyao, Y.
\newblock (2012).
\newblock {Comparing Different Criteria for Vietnamese Word Segmentation}.
\newblock In {\em Proceedings of the 3rd Workshop on South and Southeast Asian
  Natural Language Processing}, pages 53--68.

\bibitem[\protect\citename{Nguyen \bgroup et al.\egroup }2013]{Nguyen2013}
Nguyen, D.~Q., Nguyen, D.~Q., and Pham, S.~B.
\newblock (2013).
\newblock {KbQAS: A Knowledge-based QA System}.
\newblock In {\em Proceedings of the 12th International Semantic Web Conference
  (Posters \& Demonstrations Track)}, pages 109--112.

\bibitem[\protect\citename{Nguyen \bgroup et al.\egroup
  }2014]{NguyenEACL2014NPP}
Nguyen, D.~Q., Nguyen, D.~Q., Pham, D.~D., and Pham, S.~B.
\newblock (2014).
\newblock {RDRPOSTagger: A Ripple Down Rules-based Part-Of-Speech Tagger}.
\newblock In {\em Proceedings of the Demonstrations at the 14th Conference of
  the European Chapter of the Association for Computational Linguistics}, pages
  17--20.

\bibitem[\protect\citename{Nguyen \bgroup et al.\egroup
  }2016]{NguyenNPP_AICom2015}
Nguyen, D.~Q., Nguyen, D.~Q., Pham, D.~D., and Pham, S.~B.
\newblock (2016).
\newblock {A Robust Transformation-Based Learning Approach Using Ripple Down
  Rules for Part-of-Speech Tagging}.
\newblock {\em AI Communications}, 29(3):409--422.

\bibitem[\protect\citename{Nguyen \bgroup et al.\egroup }2017a]{NguyenNP_SWJ}
Nguyen, D.~Q., Nguyen, D.~Q., and Pham, S.~B.
\newblock (2017a).
\newblock {Ripple Down Rules for Question Answering}.
\newblock {\em Semantic Web}, 8(4):511--532.

\bibitem[\protect\citename{Nguyen \bgroup et al.\egroup
  }2017b]{NguyenVNDJ-ALTA-2017}
Nguyen, D.~Q., Vu, T., Nguyen, D.~Q., Dras, M., and Johnson, M.
\newblock (2017b).
\newblock {Fro{m}\ {W}ord Segmentation to POS Tagging for Vietnamese}.
\newblock In {\em Proceedings of the Australasian Language Technology
  Association Workshop 2017}, pages 108--113.

\bibitem[\protect\citename{Pham and Hoffmann}2006]{PhamSB:2006}
Pham, S.~B. and Hoffmann, A.
\newblock (2006).
\newblock {Efficient Knowledge Acquisition for Extracting Temporal Relations}.
\newblock In {\em Proceedings of the 17th European Conference on Artificial
  Intelligence}, pages 521--525.

\bibitem[\protect\citename{Pham \bgroup et al.\egroup }2009]{Pham2009}
Pham, D.~D., Tran, G.~B., and Pham, S.~B.
\newblock (2009).
\newblock {A Hybrid Approach to Vietnamese Word Segmentation using Part of
  Speech tags}.
\newblock In {\em Proceedings of the 2009 International Conference on Knowledge
  and Systems Engineering}, pages 154--161.

\bibitem[\protect\citename{Plisson \bgroup et al.\egroup }2008]{Plisson2008}
Plisson, J., Lavra\v{c}, N., Mladeni\'{c}, D., and Erjavec, T.
\newblock (2008).
\newblock {Ripple Down Rule Learning for Automated Word Lemmatisation}.
\newblock {\em AI Communications}, 21(1):15--26.

\bibitem[\protect\citename{Poowarawan}1986]{Poowarawan}
Poowarawan, Y.
\newblock (1986).
\newblock {Dictionary-based Thai Syllable Separation}.
\newblock In {\em Proceedings of the Ninth Electronics Engineering Conference},
  pages 409--418.

\bibitem[\protect\citename{Richards}2009]{RichardsD09}
Richards, D.
\newblock (2009).
\newblock {Two Decades of Ripple Down Rules Research}.
\newblock {\em Knowledge Engineering Review}, 24(2):159--184.

\bibitem[\protect\citename{Takahashi and Yamamoto}2016]{Takahashi2016}
Takahashi, K. and Yamamoto, K.
\newblock (2016).
\newblock {Fundamental tools and resource are available for Vietnamese
  analysis}.
\newblock In {\em Proceedings of the 2016 International Conference on Asian
  Language Processing}, pages 246--249.

\bibitem[\protect\citename{Thang \bgroup et al.\egroup
  }2008]{DinhQuangThang2008}
Thang, D.~Q., Phuong, L.~H., Huyen, N. T.~M., Tu, N.~C., Rossignol, M., and
  Luong, V.~X.
\newblock (2008).
\newblock {Word segmentation of Vietnamese texts : a comparison of approaches}.
\newblock In {\em Proceedings of the 6th International Conference on Language
  Resources and Evaluation}, pages 1933--1936.

\bibitem[\protect\citename{Tran \bgroup et al.\egroup }2010]{Tran2010}
Tran, T.~O., Le, A.~C., and Ha, Q.~T.
\newblock (2010).
\newblock {Improving Vietnamese Word Segmentation and POS Tagging using MEM
  with Various Kinds of Resources}.
\newblock {\em Journal of Natural Language Processing}, 17(3):41--60.

\bibitem[\protect\citename{Tran \bgroup et al.\egroup }2012]{Tran2012}
Tran, N.~A., Dao, T.~T., and Nguyen, P.~T.
\newblock (2012).
\newblock {An effective context-based method for Vietnamese-word segmentation}.
\newblock In {\em Proceedings of the 1st International Workshop on Vietnamese
  Language and Speech Processing}, pages 34--40.

\bibitem[\protect\citename{Xu and Hoffmann}2010]{Xu:2010}
Xu, H. and Hoffmann, A.
\newblock (2010).
\newblock {RDRCE: Combining Machine Learning and Knowledge Acquisition}.
\newblock In {\em Proceedings of the 11th International Workshop on Knowledge
  Management and Acquisition for Smart Systems and Services}, pages 165--179.

\end{thebibliography}

\end{document}